\documentclass{article}
\usepackage{ijcai25}

\usepackage{times}
\usepackage{soul}
\usepackage{url}
\usepackage[hidelinks]{hyperref}
\usepackage[utf8]{inputenc}
\usepackage[small]{caption}

\usepackage{newunicodechar}
\newunicodechar{∗}{\textasteriskcentered}

\usepackage{graphicx}
\usepackage{amsmath}
\usepackage{amsthm}
\usepackage{booktabs}
\usepackage{algorithm}
\usepackage{algorithmic}
\usepackage[switch]{lineno}

\usepackage[dvipsnames]{xcolor}
\usepackage{colortbl}
\usepackage{graphicx,amssymb,latexsym,amsmath,amsthm,times}

\newcommand{\best}{\cellcolor{tablered}}
\newcommand{\sbest}{\cellcolor{orange}}
\newcommand{\tbest}{\cellcolor{yellow}}

\definecolor{yellow}{rgb}{1, 1, 0.7}
\definecolor{orange}{rgb}{1, 0.85, 0.7}
\definecolor{tablered}{rgb}{1, 0.7, 0.7}
\usepackage{amsfonts}
\DeclareUnicodeCharacter{211D}{\mathbb{R}}

\usepackage{mathtools} 
\usepackage{xspace} 
\newcommand{\FLIP}{\protect\reflectbox{F}LIP\xspace}

\title{K-Buffers: A Plug-in Method for Enhancing Neural Fields \\ with Multiple Buffers}

\author{
Haofan Ren$^1$\and
Zunjie Zhu$^{1}$\footnote{∗Corresponding author. The source code is available: \url{https://github.com/renhaofan/k-buffers}}\and
Xiang Chen$^1$\and
Ming Lu$^2$\and
Rongfeng Lu$^1$\And
Chenggang Yan$^1$\\
\affiliations
$^1$Hangzhou Dianzi University\\
$^2$Intel Labs China\\
\emails
\{hfren, zunjiezhu, cchenx, rongfeng-lu, cgyan\}@hdu.edu.cn,
lu199192@gmail.com
}

\begin{document}

\maketitle

\begin{abstract}
Neural fields are now the central focus of research in 3D vision and computer graphics. Existing methods mainly focus on various scene representations, such as neural points and 3D Gaussians. However, few works have studied the rendering process to enhance the neural fields. In this work, we propose a plug-in method named K-Buffers that leverages multiple buffers to improve the rendering performance. Our method first renders K buffers from scene representations and constructs K pixel-wise feature maps. Then, We introduce a K-Feature Fusion Network (KFN) to merge the K pixel-wise feature maps. Finally, we adopt a feature decoder to generate the rendering image. We also introduce an acceleration strategy to improve rendering speed and quality. We apply our method to well-known radiance field baselines, including neural point fields and 3D Gaussian Splatting (3DGS). Extensive experiments demonstrate that our method effectively enhances the rendering performance of neural point fields and 3DGS.
\end{abstract}

\section{Introduction}

\label{sec:intro}
Since the pioneering work NeRF~\cite{mildenhall2020nerf}, neural fields have become the core research problem in 3D vision and computer graphics. Although NeRF can achieve high fidelity, it suffers from several drawbacks such as long training time and slow rendering speed. Instant-NGP~\cite{muller2022instant} uses a multiresolution hash-grid to achieve fast training and rendering speed at the expense of huge memory consumption. PlenOctrees~\cite{yu2021plenoctrees} can render $800 \times 800$ images at more than 100 FPS, but the model size has increased by nearly 100 times compared with the original NeRF. 3D Gaussian~\cite{3dgs} is the most recent representation that achieves state-of-the-art visual quality while maintaining competitive training and rendering times. However, the model size of 3D Gaussian is still much larger than the original NeRF. This presents a challenge in balancing computational cost and model size of neural fields. Conversely, neural point fields, due to their intrinsic properties, hold promise in alleviating this challenge.

\begin{figure}
\centering
    \includegraphics[width=0.5\textwidth]{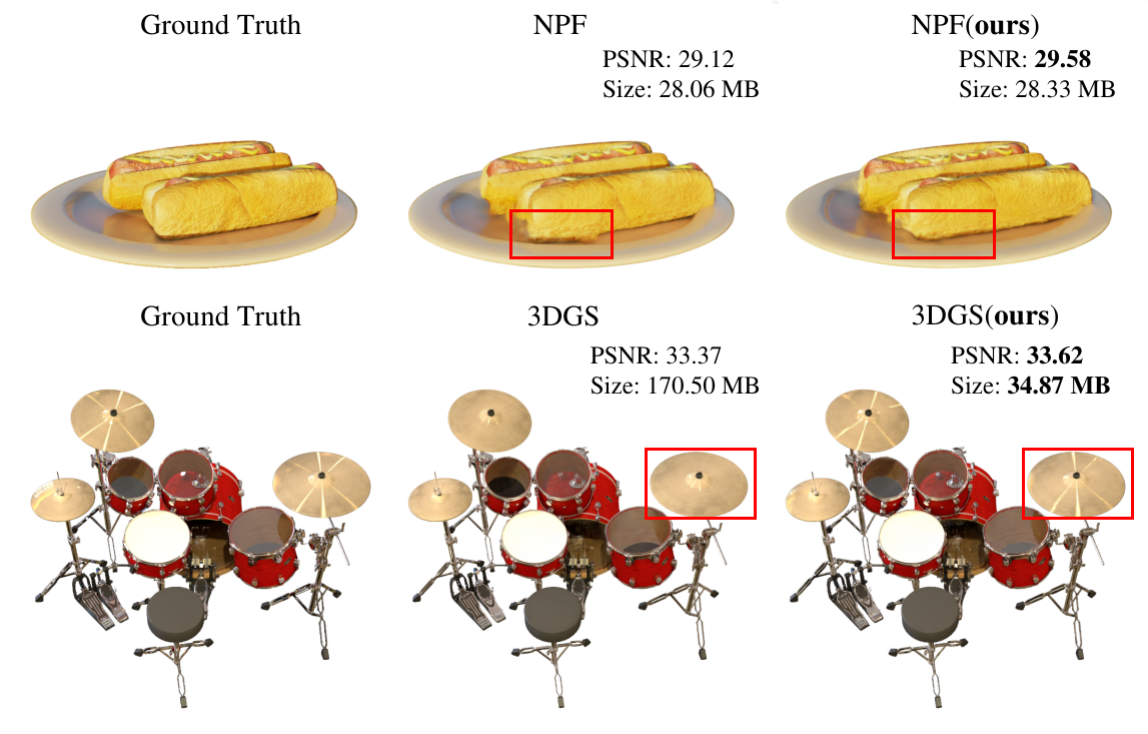}
    \caption{As shown in the figure, our method can simultaneously enhance neural point fields(NPF) and 3DGS.}
    \label{fig:teaser}
\end{figure}

\begin{figure*}[tb]
\centering
\includegraphics[width=\textwidth]{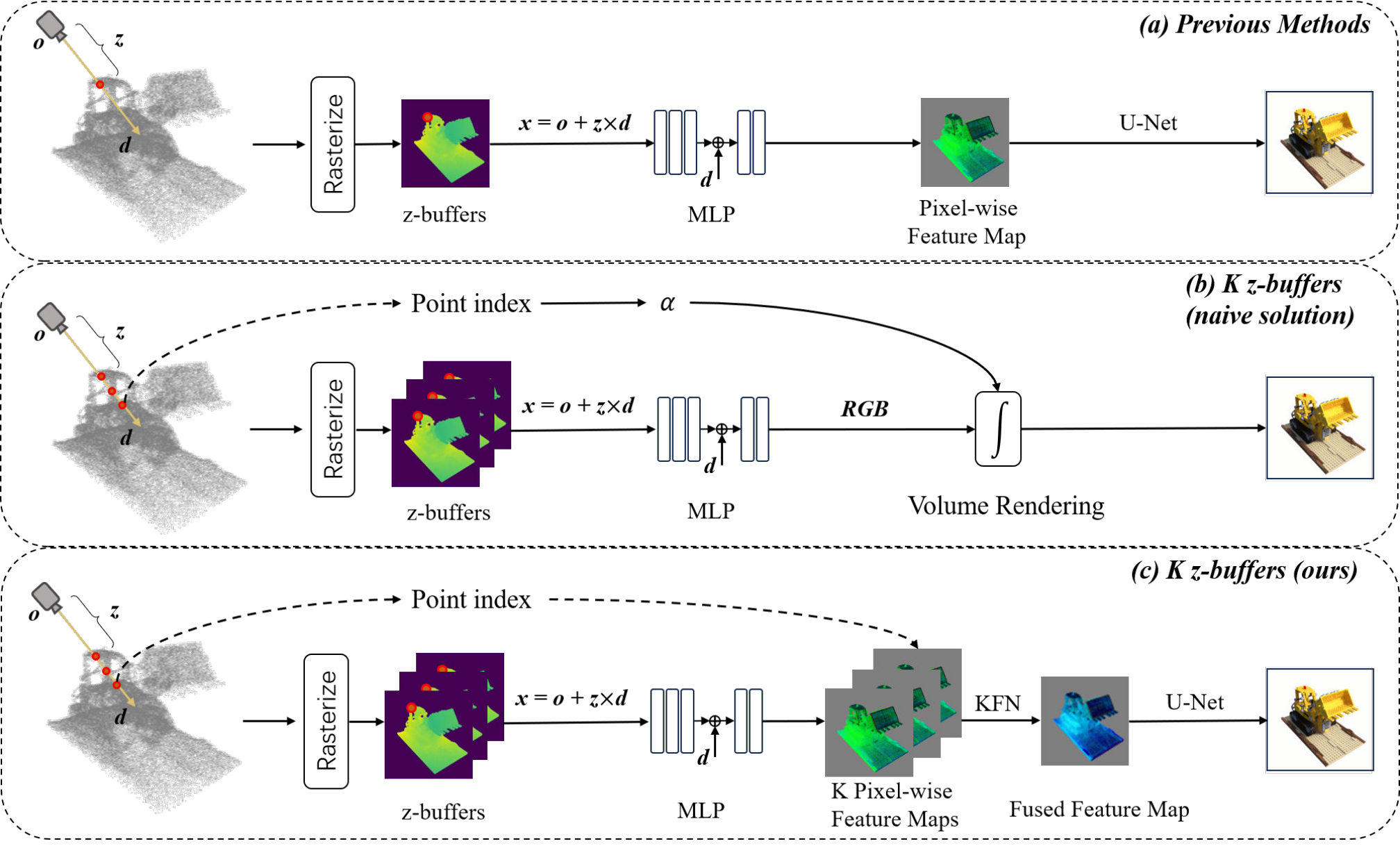}
\caption{\textbf{The motivation of our method.} (a) Previous neural point fields first render the pixel-wise z-buffers and then use a decoder to generate the image from the pixel-wise feature map. However, they are sensitive to the noisy point cloud. (b) A naive solution is to render K z-buffers and use volume rendering to integrate the K colors. However, this solution achieves overfitted results. (c) Our method uses KFN to integrate the K pixel-wise feature maps, significantly enhancing the performance of neural point fields.}
\label{fig:insight}
\end{figure*}

However, neural point fields also have some limitations. A core limitation is that they are non-trivial to render since the point clouds are noisy. Therefore, neural point fields usually fail to achieve competitive performance compared with NeRF since volume rendering is more robust than point cloud rasterization. Several recent works have been proposed to solve this limitation from different aspects. BPCR~\cite{huang2023boosting} imposes radiance mapping borrowed from NeRF~\cite{mildenhall2020nerf} to construct point features. FrePCR~\cite{zhang2023frequency} leverages a HyperNetwork~\cite{HaDL16} and narrows the positional encoding interval to improve the point features.
READ~\cite{li2023read} propose an improved U-Net~\cite{unet} named $\omega$ - net to filter the neural point descriptors on different scales to fill the holes in the rasterized image. SNP~\cite{zuo2022view} uses the spherical harmonics functions to construct point-wise features and a shallow U-Net~\cite{unet} without a normalization layer to remove noise in the feature map. However, they all ignore the core difference between volume rendering and point cloud rasterization. The key insight of volume rendering is to integrate the colors of multiple points along a ray, rather than a single point in rasterization. This motivates us to consider using multiple points to determine the rendered color of a pixel.

A naive solution is that, after obtaining K z-buffers, we can follow the approach of NeRF and use volume rendering to integrate the K colors predicted by MLPs as shown in Figure.~\ref{fig:insight}(b). Our experiments revealed that this solution suffers from severe overfitting issues. Although data augmentation methods like random crop could alleviate the overfitting issue, the rendering quality still falls short of expectations. Besides, the memory requirement increases dramatically, restricting high-resolution rendering. The main problem of this naive solution is that it simply integrates the colors along the ray, still failing to solve the noise problem of neural point fields. Instead, we propose a novel method named K-Buffers to enhance the rendering quality by integrating the K feature maps. Specifically, we first encode the K z-buffers into K feature maps and then introduce a K-Feature Fusion Network (KFN) to fuse the K feature maps in latent space. In this manner, we can reduce the noise caused by noisy neural point fields. In the process of generating the K feature maps, we introduce a strategy to improve rendering speed. Finally, we use a decoder to generate the final rendered image. Our method significantly improves the robustness against noise, whether it is inherent in the point cloud or introduced during rasterization. Furthermore, our method can also help 3DGS to improve the visual quality and reduce the model size due to its rasterization process. In summary, our contributions can be summarized as follows:
    
\begin{itemize}
    \item[$\bullet$] We propose a novel method to enhance the performance of neural point fields with multiple z-buffers.
    \item[$\bullet$] We design a tiny K-Feature Fusion Network (KFN) to reduce the noises caused by the point cloud geometry and rasterization process. 
    \item[$\bullet$] We propose a novel acceleration strategy to speed up the rendering time of the proposed method.
    \item[$\bullet$] We conduct comprehensive experiments to demonstrate the effectiveness and advantages of our method both for neural point fields and 3DGS.
\end{itemize}

\section{Related Work}
Neural Radiance Fields~\cite{mildenhall2020nerf} utilizes a 5D implicit function for scene modeling via a continuous volumetric approach, allowing it to estimate both density and radiance at any given position and direction. In this way, NeRF marked a paradigm shift in scene representation and the synthesis of realistic novel views.

\subsubsection{Accelerated Neural Radiance Fields.}
While NeRF yields remarkable results, this comes at the cost of large computation, due to the need to evaluate a large MLP model hundreds of times for each pixel. Subsequent work has proposed enhanced data structures to accelerate. Plenoxels~\cite{yu2022plenoxels} is a sparse voxel grid where each occupied voxel corner stores a scalar opacity and a vector of spherical harmonic (SH) coefficients for each color channel. PlenOctrees~\cite{yu2021plenoctrees} make use of octree and a modified NeRF that is trained to output spherical basis functions to accelerate rendering speed. InstantNGP~\cite{muller2022instant} uses multi-resolution hash grids to accelerate the training speed, reducing training time to a few seconds. KiloNeRF~\cite{reiser2021kilonerf} utilizes thousands of tiny MLPs to accelerate the rendering speed. TensorRF~\cite{chen2022tensorf} proposes to factorize the volume field tensor into multiple compact low-rank components to accelerate training speed. The above methods all use various types of data structures to store features, making them more efficiently organized in 3D space. In a sense, this is a trade-off between computation and memory.

\subsubsection{Neural Point Fields.}
Although the techniques above demonstrate remarkable effectiveness, it is difficult to adapt them to model large environments, which presents a challenge. An alternative method involves utilizing point clouds to represent the geometric structure of the scene~\cite{chang2023pointersect},~\cite{li2023read}, ~\cite{chang2023neural}. Point clouds can vary in density, allowing for the allocation of computational resources where necessary, and effectively skipping the empty space. Thus, it is not prone to generate floating artifacts, which is a server problem in neural radiance fields. 

There are two major categories of methods for neural point fields: those based on ray casting and those based on rasterization. The ray-casting-based methods achieved high rendering quality. Point-NeRF~\cite{xu2022point} and developed work~\cite{sun2023pointnerf++} usually sample hundreds or thousands of points for each ray. To perform volume rendering, the feature of each sampled point is queried in the local neighborhood of point clouds to produce density and color. NPLF~\cite{ost2022nplf} efficiently represents scenes with just one radiance evaluation per ray by promoting sparse point clouds to neural implicit light fields. However, the computational burden of these approaches remains high, making it challenging to achieve real-time rendering. HashPoint~\cite{ma2024hashpoint} proposes to accelerate ray-tracing point cloud rendering. However, even with the acceleration, it is still challenging to achieve real-time rendering. Additionally, the model sizes are several times larger than NeRF.

On the contrary, the rasterization-based methods get a great balance between model size and rendering speed. NPBG~\cite{aliev2020neural_npbg} rasterize the points with neural descriptors at different scale resolutions and a U-Net followed to obtain the rendered image. NPBG++~\cite{rakhimov2022npbg++} demonstrated that point embedding can be directly extracted from input images, allowing their method to render unseen point clouds without per-scene optimization.
ADOP~\cite{ruckert2022adop} combines neural rendering with a differentiable point rasterizer to minimize the difference between rendered images and the ground truth. Dai \textit{et al.}~\cite{dai2020neural} aggregate points into multi-plane images, combined with a 3D-convolutional network. TriVol~\cite{hu2023trivol} proposes a novel 3D representation called triple volumes to get point embedding. BPCR~\cite{huang2023boosting} combines implicit radiance mapping and rasterized z-buffers to achieve excellent results. 
The improved approach in FrePCR~\cite{zhang2023frequency} utilizes a hypernetwork to enhance radiance mapping. However, almost all rasterization-based methods do not specifically address noise introduced by rasterization, such as holes. The usual practice is to use a post-processing U-Net that involves downsampling and then upsampling the rasterized feature map. Unfortunately, such methods are not always effective. The noise mentioned above is the main reason why rasterization-based methods often have lower rendering quality compared to ray-casting-based methods. 3DGS~\cite{3dgs} utilizes anisotropic Gaussians with a geometry optimization strategy to allow a fast and high-quality radiance field. However, it still suffers from the noise issues mentioned above and significantly increases storage requirements. In this work, we focus on improving the rendering quality of rasterization-based methods without too much increase in model size and computational requirement. Additionally, we aim to enhance the robustness of these methods against noise.

\section{Method}
\begin{figure*}[tb] \centering
    \includegraphics[width=\textwidth]{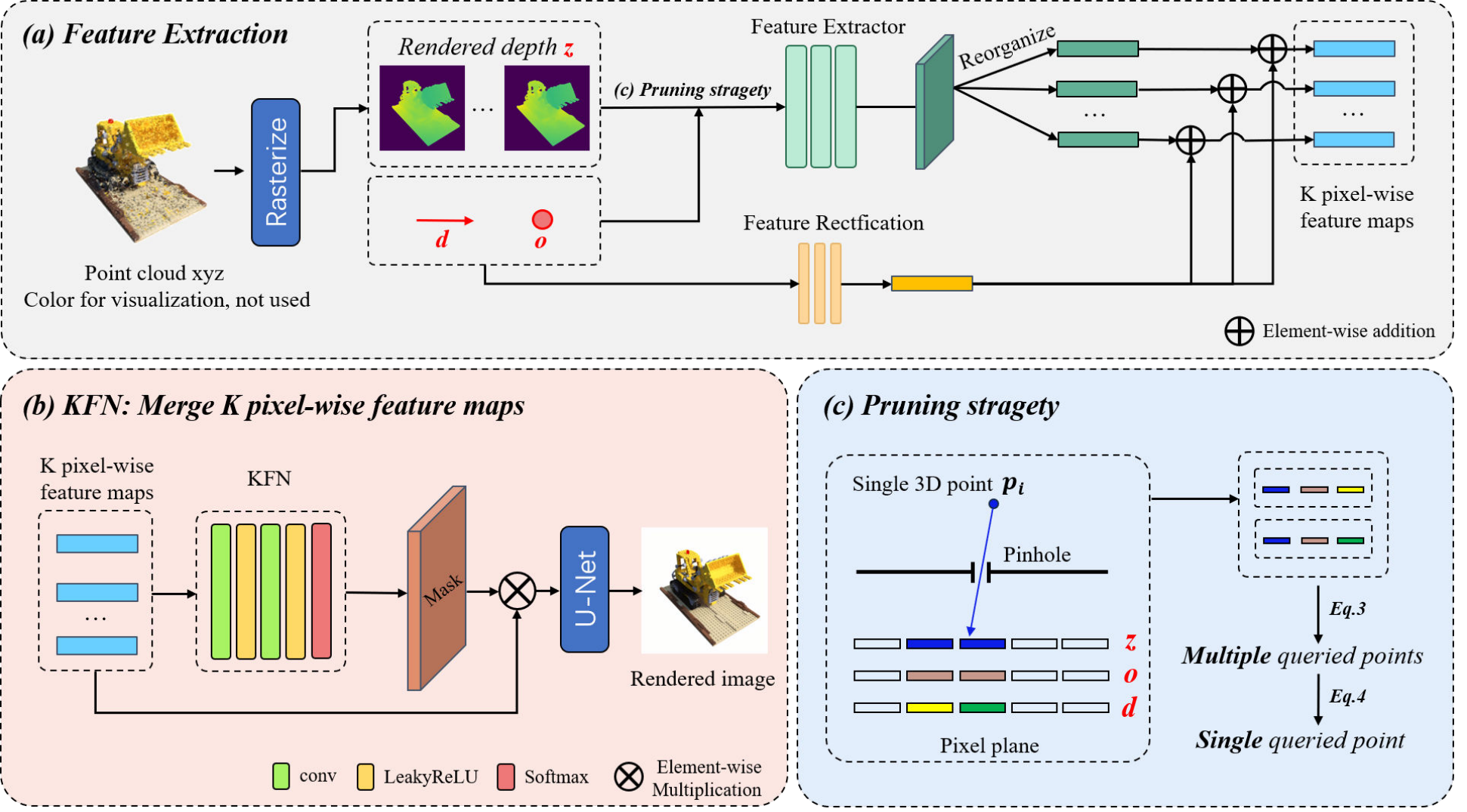}
    \caption{The Overview pipeline with our model. (a) depicts how to obtain the neural descriptors from noisy point clouds.
    (b) describe how to fuse K pixel-wise features and ultimately obtain the rendering result.
    (c) illustrates the reason why a single 3D point will generate multiple queried points and how to reduce the number of them by our pruning strategy.}
    \label{fig:pipeline}
\end{figure*}

In this paper, we propose a novel strategy to enhance the rasterization pipeline of point-based neural rendering based on K z-buffers. First, we save the K z-buffers during the depth test instead of the closest point to the camera position for each pixel. Then we encode queried points to latent space, which are the so-called neural points. However, if we directly encode all queried points of the K z-buffers, the number of neural points will increase with the growth of K. Therefore, we propose a solution to significantly reduce the number of neural points, even fewer than a single z-buffer. After obtaining the neural points, we reorganize them into K pixel-wise feature maps.
Finally, we utilize KFN to merge and denoise K feature maps and a U-Net to decode into colors. 


\subsection{Feature Extraction}%
\label{sub:Feature Extraction}

The NeRF representation~\cite{mildenhall2020nerf} takes the 3D coordinate $\boldsymbol{x} = (x, y, z)$ and view direction $\boldsymbol{d} = (\theta, \phi)$ as inputs and output the color \textbf{c} and density $\sigma$ using a Multi-Layer Perceptron (MLP) $\boldsymbol{F}_{\Theta}$, parameterized by $\Theta$:

\begin{equation}
    \label{eq:first}
    \boldsymbol{c}, \sigma=\boldsymbol{F}_{\Theta}(\boldsymbol{x}, \boldsymbol{d}).
\end{equation}

Inspired by above equation, the point-based rendering methods BPCR ~\cite{huang2023boosting}, FrePCR~\cite{zhang2023frequency} use the radiance mapping $\boldsymbol{L}$ as feature encoder:
\begin{equation}
    \label{eq:radiance mapping}
    \boldsymbol{L}=\boldsymbol{F}_{\Theta}(\boldsymbol{x}, \boldsymbol{d}).
\end{equation}
Different from NeRF, point-based rendering approaches provide explicit geometry prior. 
As a consequence, $\boldsymbol{x}$ can be acquired by Eq.~\ref{eq:ozd}:
\begin{equation}
    \label{eq:ozd}
    \boldsymbol{x} =\boldsymbol{o}+z\boldsymbol{d},
\end{equation}
where $\boldsymbol{o}$, $\boldsymbol{d}$ denote camera position and normalized ray direction respectively.
In practice, the $z$ value is obtained by querying the z-buffer via rasterization and depth test.

By recording the point closest to the camera, we can achieve a single z-buffer.
To obtain K z-buffers, we only need to store the previous K z-values during depth testing, which does not introduce excessive computational complexity compared with a single z-buffer.

The inspiration behind using multiple layers of z-buffers is that we observe different layers exhibit strengths and weaknesses in local geometry. However, due to the characteristics of rasterization, multiple layers of z-buffers as input would lead to a sharp increase in computational complexity and memory consumption. To solve this issue, we propose a novel method to prune redundant queried points.


\subsubsection{Prune redundant queried points.}
Each point is expanded to a disk of radius $\tau$ during rasterization as point clouds are discrete. 
During rasterization, a single 3D point will splash on multiple pixels' positions. These pixels have the same $z$ value and camera position $\boldsymbol{o}$, but different directions $\boldsymbol{d}_{j}$. We take a one-dimensional example to illustrate this phenomenon. As illustrated in Figure.~\ref{fig:pipeline}(c), a 3D point $\boldsymbol{p}_{i}$ of point cloud splashed onto pixel plane and occupies two pixels. These two pixels with different directions will be encoded into different queried points through Eq.~\ref{eq:ozd}. In this way, a single 3D point generates two queried points.
However, each queried point will be encoded into a neural point by Eq.~\ref{eq:radiance mapping}. As a result, the number of neural points is greater than the number of points in the point clouds.

We only reserve one direction $\boldsymbol{d}_{m}$ for each 3D point to generate the queried point. The queried points generated by other directions are referred to as \textbf{redundant queried points}.

In particular, each pixel's ID $j$ can be calculated by $ W * \text{pixel.x} + \text{pixel.y}$, 
where $W$ means the width of rendered image, $\text{pixel.x}$ and $\text{pixel.y}$ represent the indices in the pixel coordinate system respectively. For each 3D point $\boldsymbol{p}_{i}$ observed by the current view, we can deduce ID collection $\boldsymbol{A}_{\boldsymbol{p}_{i}}$ of all pixels occupied by $\boldsymbol{p}_{i}$. We only keep the direction $\boldsymbol{d}_{m}$ corresponding to the smallest pixel ID to generate queried point:
\begin{equation}
    \label{eq:prune}
    \boldsymbol{d}_{m} = \boldsymbol{d}_{j}, j =\min{\boldsymbol{A}_{\boldsymbol{p}_{i}}}
\end{equation}

In the example shown in Figure.~\ref{fig:pipeline}(c), five pixels' IDs are 0, 1, 2, 3, and 4, respectively. Thus, $\boldsymbol{A}_{\boldsymbol{p}_{i}}=\{1,2\}$ and $\boldsymbol{d}_{m} = \boldsymbol{d}_{1}$.








\subsubsection{Feature rectification.}

Simply pruning the queried point using Eq.~\ref{eq:prune} leads to identical features across different pixel positions occupied by $\boldsymbol{p}_{i}$, which diminishes the sensitivity of radiance mapping to variations in pixel positions.

To address this problem, we utilize a tiny MLP $\boldsymbol{T}_{\Psi}$ with 3,656 trainable params parameterized by $\Psi$ to rectify the features of queried points via Eq.~\ref{eq:radiance mapping}. 
Thus, our rectified feature map is formulated as Eq.~\ref{eq:rect_rdm}. 
Besides, we make use of different encoding methods for $\boldsymbol{F}_{\Theta}$ 
and $\boldsymbol{T}_{\Psi}$.
In practice, we apply position encoding introduced in NeRF~\cite{mildenhall2020nerf} to $\boldsymbol{x}_{m}$ and $\boldsymbol{d}_{m}$, while employing sphere harmonics encoding for $\boldsymbol{d}_{j}$. The multi-resolution hash encoding is applied to $\boldsymbol{o}$. Our rectified features can be formulated by Eq.~\ref{eq:rect_rdm}:

\begin{equation}
    \label{eq:rect_rdm}
    \boldsymbol{L}_{rect} = \boldsymbol{F}_{\Theta}(\boldsymbol{x}_{m}, \boldsymbol{d}_{m}) + \boldsymbol{T}_{\Psi}(\boldsymbol{o}, \boldsymbol{d}_{j}).
\end{equation}

\subsection{K-Feature Fusion Network}%
\label{sub:K-Feature Fusion Network}

Due to multiple z-buffers as input, the key is how to fuse and denoise them. A similar research area~\cite{godard2018deep},~\cite{mildenhall2018kpn},~\cite{xia2020bpn} is called burst denoise. It takes a burst of noisy images captured with a handheld camera as input, and the aim is to produce a clean image. KPN~\cite{mildenhall2018kpn}(Kernel Prediction Network) is a kind of classic and widely used technique in this area. KPN predicts a feature vector for each pixel, which is then reshaped into a set of spatially varying kernels applied to the input burst images. A naive idea is to direly employ KPN~\cite{mildenhall2018kpn} to process multiple z-buffers and merge them into a single clean z-buffer, which is then fed into the point-based rendering methods~\cite{huang2023boosting},~\cite{zhang2023frequency}. Nevertheless, this approach can easily lead to unstable training scenarios.


We designed a tiny network to integrate multiple K z-buffers in the latent space. In other words, we merge K pixel-wise feature maps and denoise to generate cleaner feature maps, where the denoise operation happens in latent space rather than the aspect of the z-buffer.
In practice, we first generate K pixel-wise feature maps from K z-buffers. Then, we propose a module called \textbf{K}-Feature \textbf{F}usion \textbf{N}etwork(KFN) to fuse K pixel-wise feature maps.
Different from the U-Net structure of KPN, KFN is very simple and consists of two convolution layers, two activation layers, and a softmax normalization(see Figure.~\ref{fig:pipeline}(b)). Besides, KFN predicts a pixel-wise scalar mask range in $[0, 1]$ to merge K pixel-wise feature maps rather than the spatially varying kernels. Besides, previous work~\cite{hedman2018deep} also uses CNNs to estimate blending weights for novel-view synthesis, which results in temporal flickering. However, our method does not suffer such a problem(see (Table~\ref{tab:ab_tc})). The results also demonstrate the good denoising effects with KFN(see (Figure.~\ref{fig:z-defect})).

\subsection{Overall Training Process}%
\label{sub:Training Strategy}
\subsubsection{Neural Points Fields.}

Our pipeline could be divided into five stages: rasterization, neural points construction, pruning, K pixel-wise feature maps fusion, and refinement. Firstly, through rasterization, all points $\boldsymbol{P}$ observed by the current view will be splashed into the pixel plane to generate K z-buffers. Since this process does not require differentiability, it can be implemented using hardware-accelerated frameworks such as OpenGL~\cite{shreiner2009opengl} and run in real-time. At the same time, we record the pixel positions $\boldsymbol{A}_{\boldsymbol{p}_{i}}$ occupied by each unique point $\boldsymbol{p}_{i} \in \boldsymbol{P}$ for reorganization.

For each pixel, we generate the queried points $\boldsymbol{x}$ from K z-buffers by Eq.~\ref{eq:ozd}. Afterward, we remove redundant points by Eq.~\ref{eq:prune}, and the remaining points $\boldsymbol{x_{m}}$ are passed through Eq.~\ref{eq:rect_rdm} to obtain latent features $\boldsymbol{L}_{rect} = N \times C$, where $N$ and $C$  refer to the number of queried points $\boldsymbol{x}_{m}$ and the dimension of latent features respectively. Using the pre-reserved set $\boldsymbol{A}_{\boldsymbol{p}_{i}}$, we reorganize the $\boldsymbol{L}_{rect}$ back to K pixel-wise feature maps with shape $K \times H \times W \times C$, while we assign zero to the feature of those unoccupied pixels. We use KFN to predict a mask with the same dimensions as K pixel-wise feature maps. The fused features would be obtained by the element-wise multiplication between the predicted mask and K pixel-wise feature maps. Finally, a U-Net is employed to decode the fused features into the rendered image in the shape of $H \times W \times 3$.

\subsubsection{3DGS.}
Each of the 3D Gaussians is parametrized by its potision $\boldsymbol{\mu}$, covariance matrix $\boldsymbol{\Sigma}$, opacity $\boldsymbol{\alpha}$ and high dimension feature $\boldsymbol{f}$. We use a tiny MLP $\boldsymbol{H}_{\Gamma}$ parameterized by ${\Gamma}$ to generate the radiance mapping $\boldsymbol{L}_{rect}$, which is constructed by blending the multiple ordered Gaussians as:

\begin{equation}
\boldsymbol{L}_{rect}=\sum_{i=1}^N \alpha_i \prod_{j=1}^{i-1}\left(1-\alpha_j\right) \boldsymbol{l}_{i}^{rect}
\label{eq:3dgs_rd}
\end{equation}

\begin{equation}
    \boldsymbol{l}_{i}^{rect} = \boldsymbol{H}_{\Gamma}(\boldsymbol{d}_{i}) + \boldsymbol{f}_{i}
\end{equation}
where $\boldsymbol{f}_{i}$, $\alpha_{i}$ and $\boldsymbol{d}_{i}$ represent the feature, opacity and view direction of queried Gaussian $i$.

In summary, for neural point fields, we incorporate the pruning queried points strategy and feature rectification into the radiance mapping $\boldsymbol{L}_{rect}$, which will be reorganized into K pixel-wise feature maps. As for 3DGS, the radiance mapping is generated by Eq.~\ref{eq:3dgs_rd}. With the radiance mapping, a proposed lightweight module named KFN is used to merge K pixel-wise feature maps. Finally, similar to the previous method, a U-Net is used to decode and obtain the rendered image.
The loss function remains completely consistent with the baseline without any modifications.

\subsection{Benchmark Evaluation}%
\label{sub:Experimental Settings}

\begin{figure}
\centering
    \includegraphics[width=0.5\textwidth]{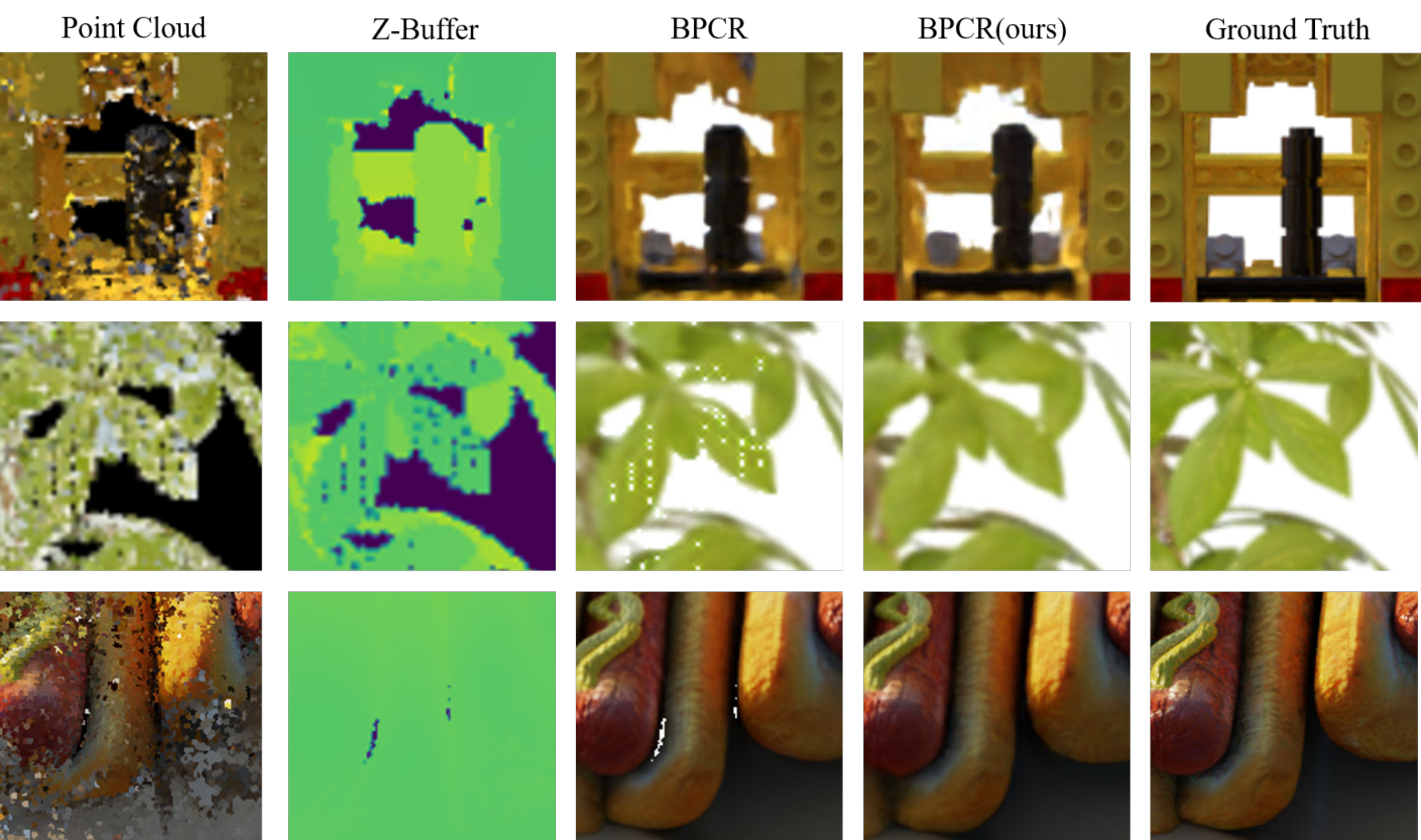}
    \caption{Illustration of the z-buffer defect. The color of the point cloud has not been used, which is only for visualization purposes.}
    \label{fig:z-defect}
\end{figure}

\begin{figure*}[h!]
\begin{center}
  \includegraphics[width=\linewidth]{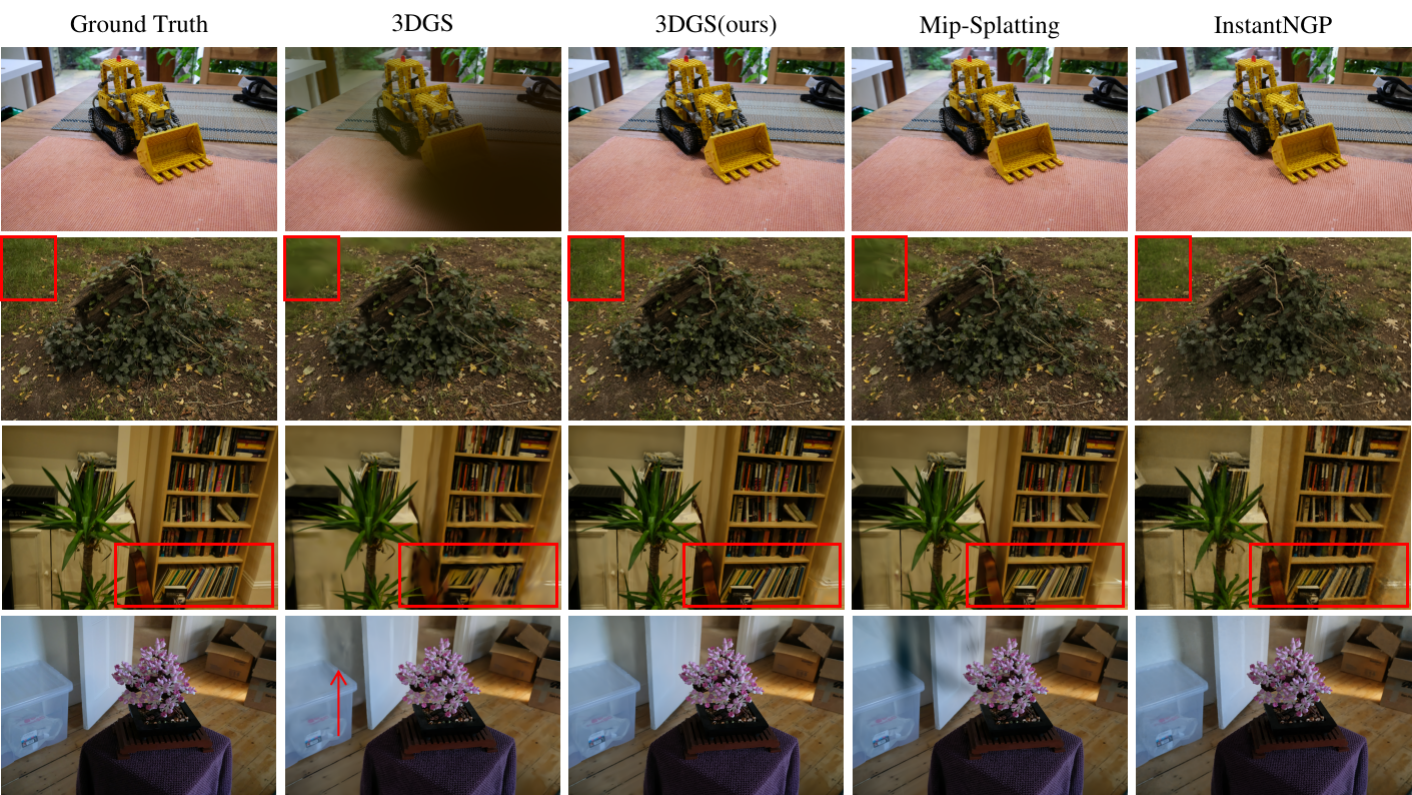}
\end{center}
\caption{We compare the rendered novel views between ours and previous methods. The scenes are, from the top down: \textit{kitchen}, \textit{stump}, \textit{room}, \textit{bonsai} from Mip-NeRF 360 dataset.
}
\label{fig:mipnerf}
\end{figure*}

\subsubsection{Settings and Compared Methods.}
Since our work needs z-buffers rasterized by point cloud, we choose three methods as our baseline to evaluate: 1) BPCR~\cite{huang2023boosting}: A simple but effective architecture point-based rendering method. 2) FrePCR~\cite{zhang2023frequency}: The improved version of BPCR. For a fair comparison, we don't narrow the positional encoding interval as described in FrePCR but rather keep it the same as BPCR.
3) 3DGS~\cite{3dgs}: An efficient point-based rendering method, which represents the scene with 3D Gaussians to optimize the geometry. We keep $\boldsymbol{F}_{\Theta}$ and U-Net exactly the same as the baseline for experimentation. We evaluate on three datasets: NeRF-Synthetic~\cite{mildenhall2020nerf}, ScanNet~\cite{dai2017scannet}, DTU~\cite{jensen2014dtu}. The point cloud initialization is same as the BPCR. Besides, we also evaluate the 3DGS on Mip-NeRF 360~\cite{barron2022mipnerf360}, which is compared with InstantNGP~\cite{muller2022instant} and Mip-Splatting~\cite{Yu2024MipSplatting}.


\begin{table}[tb]\centering
    \resizebox{0.48\textwidth}{!}{
    \large
    \begin{tabular}{*{10}{c}}
        \toprule
        Metric & PSNR $\uparrow$ & SSIM $\uparrow$ & LPIPS $\downarrow$ & Storage(MB) $\downarrow$ \\
        \midrule
        InstantNGP & 26.67 & 0.738 & 0.369 & 38.57 \\
        Mip-Splatting & 29.02 & 0.871 & 0.224 & 768.86 \\
        \midrule
        3DGS & 28.93 & \textbf{0.869} & 0.136 & 735.57 \\
        3DGS+Ours& \textbf{29.19} & 0.859 & \textbf{0.126} & \textbf{383.43} \\
        \bottomrule
    \end{tabular}
    }
    \caption{Quantitative comparisons on Mip-NeRF 360 dataset.}
    \label{tab:mipnerf360}
\end{table}

\subsubsection{Quantitative and Qualitative Results.}
\begin{figure}[htbp]
\centering
\includegraphics[width=\linewidth]{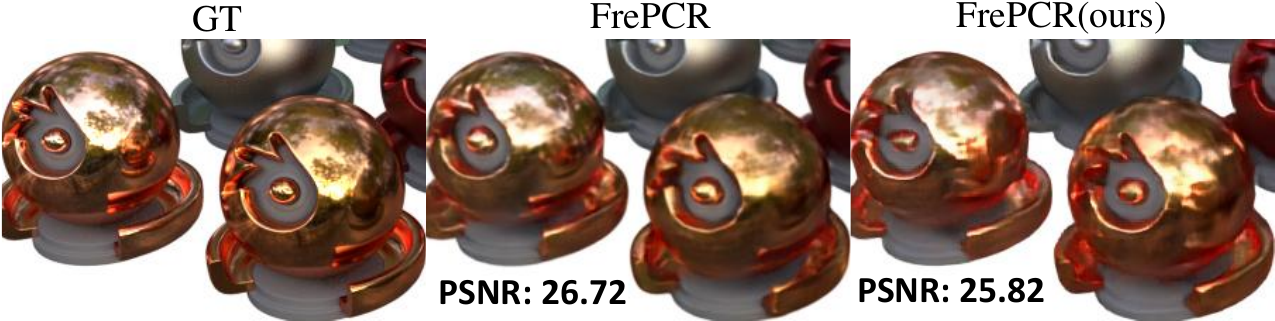}
\caption{Our method struggles to produce accurate results in regions with specular reflections.}
\label{fig:failure_cases}
\end{figure}

We present the quantitative results in Table~\ref{tab:mipnerf360} and Table~\ref{tab:benchmark}. It can be found that our method can simultaneously help improve the rendering quality of neural point fields and 3DGS.
As shown in Figure~\ref{fig:z-defect}, the first row showcases the noise removal capability of our proposed method. The second row demonstrates the aliasing caused by unsuitable point radius and camera observation view, which can be alleviated by our method. The third row exhibits its capability for filling holes. As illustrated in Figure~\ref{fig:mipnerf}, our method can effectively help 3DGS mitigate the floater artifacts. At the same time, it can restore high-frequency details of the scene and also improve modeling in low-textured areas. Besides, we noticed that the PSNR of FrePCR on the DTU dataset is bad shown in Table~\ref{tab:benchmark}. One possible reason is that the scenes of DTU contain relatively less high-frequency information. FrePCR increases the expression ability of neural points by AFNet(Adaptive Frequency Net). However, due to the low-frequency information of the scene, it is difficult to modulate the implicit radiance signal correctly for AFNet. Incorporating our method is equivalent to expanding the receptive field of AFNet, which enhances the comprehensive understanding of the scene. We compare with the ray-tracing-based method as shown in Table~\ref{tab:pbrt}, which demonstrates the advantages and disadvantages of our method. 
We visualize the failure case of our method. As shown in Figure~\ref{fig:failure_cases}, our method blurs the details of the reflective regions.


\begin{table*}[!ht]
    \centering
    \begin{tabular*}{\hsize}{@{}@{\extracolsep{\fill}}lccccccccc@{}}
    \toprule
     & \multicolumn{3}{c}{NeRF-Synthetic} & \multicolumn{3}{c}{ScanNet }  & \multicolumn{3}{c}{DTU}  \\
     \text{} & 
     PSNR$\uparrow$ & SSIM$\uparrow$ & LPIPS$\downarrow$ & PSNR$\uparrow$ & SSIM$\uparrow$ & LPIPS$\downarrow$ & PSNR$\uparrow$ & SSIM$\uparrow$ & LPIPS$\downarrow$ \\
    \midrule
    BPCR &
    29.12 & 0.935 & 0.056 &
    25.88 & \textbf{0.794} & 0.415 &
    30.81 & 0.904 & \textbf{0.128} \\
    BPCR+Ours &
    \textbf{29.58} & \textbf{0.936} & \textbf{0.054} &
    \textbf{26.66} & 0.789 & \textbf{0.383} &
    \textbf{30.99} & \textbf{0.907} & 0.151 \\
    \midrule
    FrePCR &
    29.26 & 0.934 & 0.055 &
    \textbf{26.44} & \textbf{0.788} & \textbf{0.384} &
    24.61 & 0.870 & 0.195 \\
    FrePCR+Ours &
    \textbf{29.83} & \textbf{0.939} & \textbf{0.051} &
    26.34 & \textbf{0.788} & 0.391 &
    \textbf{30.61} & \textbf{0.905} & \textbf{0.153} \\
    \midrule
    3DGS &
    33.37 & 0.968 & 0.028 &
    24.16 & 0.773 & \textbf{0.403} &
    \textbf{33.91} & 0.943 & \textbf{0.050} \\
    3DGS+Ours &
    \textbf{33.62} & \textbf{0.969} & \textbf{0.019} &
    \textbf{25.55} & \textbf{0.785} & 0.406 &
    \textbf{33.91} & \textbf{0.954} & 0.059 \\
    \bottomrule
    \end{tabular*}
    \caption{Quantitative comparisons on NeRF-Synthetic, ScanNet and DTU dataset.}
    \label{tab:benchmark}
\end{table*}

\begin{table}[tb]\centering
    \resizebox{0.48\textwidth}{!}{
    \label{tab:point_rr}
    \large
    \begin{tabular}{l|cccccc|cc}
    \toprule
    & \multicolumn{6}{c|}{\text{Point-based (Rasterization)}} & \multicolumn{2}{c}{\text{Point-based (Ray tracing)}} \\
    & BPCR & BPCR+Ours & FrePCR & FrePCR+Ours & 3DGS & 3DGS+Ours & PN & PN+HP \\
    \midrule
    PSNR$\uparrow$ & 29.12 & 29.58 & 29.26 & 29.83 & \sbest33.37 & \best33.62 & \tbest33.31 & 33.22 \\
    FPS$\uparrow$ & 39.21 & 27.52 & \tbest39.56 & 26.39 & \best294.70 & \sbest82.12 & 0.12 & 9.60 \\
    Storage(MB)$\downarrow$ & \best28.46 & \sbest28.73 & 28.63 & 28.90 & 170.50 & \tbest34.87 & 105.12	& 105.12 \\
    \bottomrule
    \end{tabular}
    }
    \caption{Comparison of our method integrated with ray-tracing-based method PN(PointNeRF) and HP(HashPoint). The storage includes the point cloud size.}
    \label{tab:pbrt}
\end{table}

\begin{table}[tb]\centering
    \resizebox{0.48\textwidth}{!}{
    \large
    \begin{tabular}{lcccccc}
    \toprule
    \text{} & Number & GFLOPS & Model Size(MB) & Mem(GB) & PSNR & FPS \\
    \midrule
    BPCR & 203,757 & \textbf{80.96} & \textbf{8.20} & 9 & 29.12 & \textbf{39.21}\\
    BPCR(KFN) & 1,613,013 & 386.86 & 8.46 & 33 & 29.95 & 10.12 \\
    BPCR(KFN+Pruning) & \textbf{194,225} & 122.43 & 8.46 & \textbf{7} & 29.68 & 27.80 \\
    BPCR(KFN+Pruning+Rect) & \textbf{194,225} & 122.44 & 8.47 & \textbf{7} & \textbf{29.58} & 27.52 \\
    \bottomrule
    \end{tabular}
    }
    \caption{Complexity analysis on NeRF-Synthetic dataset. The first column of metrics means the number of queried points $\boldsymbol{x}$. Since the numbers and GFLOPS are view-dependent, here we take the first view of the test scene as an example.}
\label{tab:analysis}
\end{table}

\subsection{Ablation Study}%
As shown in Table~\ref{tab:analysis}, our proposed pruning strategy significantly reduces the number of queried points, leading to a substantial decrease in memory usage and computational load. Through the pruning strategy, both computational and memory usage have been reduced by more than 3 times. It seems that feature rectification does not improve rendering quality. However, we conducted experiments on different datasets as shown in Table~\ref{tab:ab_module}. We find that feature rectification achieves the best overall rendering quality across multiple datasets. In addition, each rendered pixel is not derived from the fixed number of Gaussians in 3DGS process, which hinders the efficient implementation of the pruning operation on CUDA. As a result, the pruning operation is not applied to 3DGS in our approach.

We use the StopThePop~\cite{radl2024stopthepop} to evaluate the temporal consistency, utilizing MSE(Mean Squared Error) and \FLIP~\cite{Andersson2020} as the evaluation metrics. As shown in Table~\ref{tab:ab_tc}, integrating our method does not compromise the temporal consistency of the original approach.

We conduct tests to evaluate the impact of varying z-buffer layers as shown in Table~\ref{tab:ab_k}. It is clear that as K increases, there is an increase in FPS, model size and rendering quality. The model size does not sharply increase with the increment in K, but the computational load becomes more sensitive to K.



\begin{table}[htbp]
    \centering
    \begin{tabular}{lcc|cc}
        \toprule
        & \multicolumn{2}{c|}{\text{Short-range}} & \multicolumn{2}{c}{\text{Long-range}} \\
        \text{} & MSE$\downarrow$ & \FLIP$\downarrow$ & MSE$\downarrow$ & \FLIP$\downarrow$ \\
        \midrule
        3DGS     & 0.018 & 0.008 & 0.100 & 0.028 \\
        3DGS+Ours  & 0.018 & 0.009 & 0.102 & 0.028 \\
        \bottomrule
    \end{tabular}
    \caption{Quantitative results of temporal consistency on NeRF-Synthetic dataset. We use step 1,7 as frame offset during optical flow prediction to evaluate short-range and long-range consistency, respectively.}
\label{tab:ab_tc}
\end{table}

\begin{table}[htbp]
    \centering
    \begin{tabular}{lcccc}
    \toprule
    \text{} & K=1 & K=2 & K=4 & K=8 \\
    \midrule
    PSNR$\uparrow$ & 26.17 & 26.29 & 26.84 & \textbf{27.62} \\
    FPS$\downarrow$ & \textbf{51.00} & 44.87 & 36.83 & 27.52 \\
    Model Size(MB)$\uparrow$ & \textbf{8.30} & 8.33 & 8.37 & 8.47 \\
    \bottomrule
    \end{tabular}
    \caption{Quantitative results on \textit{lego} scene of NeRF-Synthetic for different values of K.}
\label{tab:ab_k}
\end{table}

\begin{table}[H]\centering
    \resizebox{0.48\textwidth}{!}{
    \large
    \begin{tabular}{lcccc}
    \toprule
    \text{} & NeRF-Synthetic & ScanNet & DTU & Mean\\
    \midrule
    BPCR & 29.12 & 25.88 & 30.81 & 28.60  \\
    BPCR(KFN) & \textbf{29.95} & 25.93 & 31.20 & 29.03  \\
    BPCR(KFN+Pruning) & 29.68 & 26.05 & \textbf{31.08} & 28.94  \\ 
    BPCR(KFN+Pruning+Rect) & 29.58 & \textbf{26.66} & 30.99 & \textbf{29.08}  \\
    \midrule
    FrePCR & 29.26 & \textbf{26.44} & 24.61 & 26.77  \\
    FrePCR(KFN) & 29.44 & 26.20 & 29.20 & 28.28  \\
    FrePCR(KFN+Pruning) & 29.68 & 26.12 & 30.78 & 28.86  \\ 
    FrePCR(KFN+Pruning+Rect) & \textbf{29.83} & 26.34 & \textbf{30.61} & \textbf{28.93}  \\
    \midrule
    3DGS & 33.37 & 24.16 & \textbf{33.91} & 30.48  \\
    3DGS(KFN) & 33.20 & \textbf{25.81} & 33.66 & 30.89  \\
    3DGS(KFN+Pruning) & / & / & / & / \\
    3DGS(KFN+Rect) & \textbf{33.62} & 25.55 & \textbf{33.91} & \textbf{31.03} \\
    \bottomrule
    \end{tabular}
    }
    \caption{The impact of different modules on PSNR across multiple datasets.}
    \label{tab:ab_module}
\end{table}

\section{Conclusion}
In this work, we propose a point cloud rendering method to enhance the neural point fields and 3DGS. We propose to utilize multiple points rather than single surface points to decide the color for each pixel. We also propose a pruning strategy and feature rectification to reduce the number of neural points. Besides, our proposed method can help 3DGS improve rendering quality and reduce storage. Experiments on major benchmarks have demonstrated the effectiveness of our method.



\appendix

\section{Benchmark Details}

\subsection{NeRF-Synthetic}
NeRF-Synthetic~\cite{mildenhall2020nerf} comprises a high-quality synthetic dataset featuring pathtraced images of 8 objects with intricate geometry and authentic non-Lambertian materials. Each object includes 100 training images and 200 testing images. The point cloud we used is provided by Point-NeRF~\cite{xu2022point}. During the training stage, data augmentation is performed through random scaling and cropping, following methodologies established in previous studies BPCR~\cite{huang2023boosting} and FrePCR~\cite{zhang2023frequency}. The training window size is fixed at $800 \times 800$. During testing, novel view images are rendered at the original resolution of $800 \times 800$.

\subsection{ScanNet}
ScanNet~\cite{dai2017scannet} is a large RGBD indoor dataset. We assess our approach on three scenes, namely \textit{scene0000\_00}, \textit{scene0043\_00} and \textit{scene0045\_00}, following the train-test split methodology employed in NPBG++. Specifically, for scenes containing less than 2000 frames, we select 100 frames with uniform spacing in the image stream. For scenes with more than 2000 frames, we select every 20th image to ensure that the training views are not overly sparse. \textit{scene0000\_00} falls into the latter category, while the other two belong to the former. Subsequently, we sample 10 frames at regular intervals for testing, allocating the remainder for training purposes. The training window size is defined as $720 \times 720$, while the test resolution is set at $960 \times 1200$.

\subsection{DTU}
DTU~\cite{jensen2014dtu} comprises multi-view stereo data captured at a resolution of 1200 × 1600. To validate our approach, we focus on three scenes scan110, scan114 and scan118, following the methodology of NPBG++, and utilize the same point clouds as theirs, obtained via PatchMatchNet~\cite{wang2020patchmatchnet}. We adopt the identical train-test split within each scene. In NPBG++, the evaluation encompasses the entire image, including the background. However, based on our empirical observations, we note that the background tends to be dark and distant from the camera, resulting in an excessively noisy point cloud. Although the models excel in rendering high-quality object details, the numerical results remain low. We contend that such evaluation settings do not align with our visual effects and are not reasonable. Therefore, to address this, we mask out the background both during training and testing using the binary segmentation masks provided by IDR~\cite{yariv2020multiview}. All methods undergo evaluation under these settings. The training window size is configured as $800 \times 800$, and the test resolution remains at $1200 \times 1600$.

\subsection{Mip-NeRF 360}
Mip-NeRF 360~\cite{barron2022mipnerf360} consists of a collection of indoor and outdoor object-centric scenes. The camera trajectory follows an orbit around the object, maintaining a fixed elevation and radius. We assess our approach on 7 scenes, namely \textit{bicycle}, \textit{bonsai}, \textit{counter}, \textit{garden}, \textit{kitchen}, \textit{room}, \textit{stump}. The point cloud initialization is from COLMAP~\cite{schoenberger2016sfm}.

\section{Implementation Details}
\subsection{Encoding Interval}
For neural point fields, we choose two baselines as our feature extractor, BPCR and FrePCR.
We follow the position encoding form of NeRF instead of narrowing the encoding interval as done by FrePCR.
\begin{equation}
\begin{aligned}
\gamma(\mathbf{p})= & \left(\sin \left(2^0 \pi \mathbf{p}\right), \cos \left(2^0 \pi \mathbf{p}\right)\right. \\
& \sin \left(2^{1} \pi \mathbf{p}\right), \cos \left(2^{1} \pi \mathbf{p}\right) \\
& \cdots, \\
& \left.\sin \left(2^{L-1} \pi \mathbf{p}\right), \cos \left(2^{L-1} \pi \mathbf{p}\right)\right)
\end{aligned}
\end{equation}

We set $L$ as 10 and 4 for spatial coordinates and view directions, respectively. Thus, BPCR and FrePCR feature extractors have the same feature dimension. The input feature dimension is 60, and the hidden dimensions are 256, 256, 256 and 128. The view directions are concatenated with the output of the second fully connected layer, which remains identical for BPCR and FrePCR.


\subsection{Hyperparameter}
We set the point cloud radius $\tau$ as 5e-3 for NeRF-Synthetic, 1.5e-2 for ScanNet, and 3e-3 for DTU. There on two ways of rasterization: OpenGL~\cite{shreiner2009opengl} for real-time rendering and PyTorch3D~\cite{ravi2020pytorch3d} for headless server. Since our training takes place on a headless server, we opt for the latter approach.
To speed up the training process, we only rasterize the point clouds once and save the fragments for reuse, since the training and testing views are fixed. To optimize memory utilization, we exclusively transfer the z-buffer corresponding to each camera view from the CPU to the GPU solely during network input. We set both the layers of z-buffer and the feature dimension to 8. In the implementation of BPCR, the output feature dimensions of each MLP $\boldsymbol{F}_{\Theta}$ layer are 256, 256, 256, 128, and 8 respectively. The ray directions $\boldsymbol{d}$ are concatenated with the output of the second fully connected layer. As for FrePCR, we don't narrow the position encoding interval. Thus, except for the hypernetwork, the output feature dimensions are the same as BPCR. We set the feature dimension to $C=8$ for neural point fields and $C=32$ for 3DGS, respectively.

For neural point fields, the output feature dimensions of MLP $\boldsymbol{T}_{\Psi}$ layer are 64 and 8 respectively. The U-Net architecture we used is the same as BPCR, which utilizes gated blocks comprising gated convolution, ReLU, and instance normalization layers. The U-Net reduces the size of the feature map four times through down-sampling. The feature dimensions of each layer are as follows: 16, 32, 64, 128 and 256. We use the Adam optimizer for training, with a batch size of 1. The initial learning rates of $\boldsymbol{F}_{\Theta}$, $\boldsymbol{T}_{\Psi}$, KFN and U-Net are 5e-4, 1e-4, 1.5e-4 and 1.5e-4 respectively, which are multiplied by 0.9999 in each step.

As for 3DGS, we used gsplat~\cite{gsplat} as the codebase due to its ease of use. 
The feature dimensions of each layer $\boldsymbol{H}_{\Gamma}$ are as follows: 64, 64 and 32. For the Mip-NeRF 360, we set the number of iterations to 60,000, while for the other three datasets, it was set to 30,000. The initial learning rates of $\boldsymbol{H}_{\Gamma}$, KFN and U-Net are 1e-3, 1.5e-4, and 1.5e-4 respectively. Other hyperparameters remain the same as the original 3DGS.

\begin{table}[tb]\centering
    \caption{The PSNR results with different encoding for feature rectification on BPCR. The results in the table correspond to a single scene from the dataset, namely \textit{lego}, \textit{scene0000\_00}, and \textit{scan110}, respectively.}
    \resizebox{0.48\textwidth}{!}{
    \large
    \begin{tabular}{*{10}{c}}
        \toprule
        Encoding &  NeRF-Synthetic & ScanNet & DTU & Mean \\
        \midrule
        o+d	& 27.52	& 24.39	& 30.46	& 27.46 \\
        Hash(o+d) & \textbf{27.80} & 24.27 & 28.62 & 26.89 \\
        SH(o+d) & 27.58 & 24.37 & \textbf{30.66} & 27.53 \\
        Hash(o)+SH(d) & 27.62 & \textbf{24.58} & 30.63 & \textbf{27.61} \\
        \bottomrule
    \end{tabular}}
    \label{table:encoding}
\end{table}

\begin{table}[tb]\centering
    \caption{Quantitative comparison of different choice for $d_{m}$ on \textit{lego} scene of NeRF-Synthetic dataset on BPCR.}
    \resizebox{0.48\textwidth}{!}{
    \large
    \begin{tabular}{*{10}{c}}
        \toprule
        \text{} &  Minimum & Random & Average \\
        \midrule
        PSNR$\uparrow$ & 27.62 & 27.63 & 27.60 \\
        SSIM$\uparrow$ & 0.915 & 0.916 & 0.913 \\ 
        LPIPS$\downarrow$ & 0.076 & 0.077 & 0.074 \\
        \bottomrule
    \end{tabular}}
    \label{table:dm}
\end{table}

\section{More Detailed Results}
As illustrated in Table~\ref{table:encoding}, we explore different encoding combinations for camera origin and direction in feature rectification. 

Acceleration strategy choose the minimum of $\boldsymbol{A}_{\boldsymbol{p}_{i}}$ as $\boldsymbol{d}_{m}$ as describe in Eq.~\ref{eq:prune}. We also test the different choices of $d_{m}$ as shown in Table~\ref{table:dm}. Each point will occupy multiple pixel positions. \textit{Random} means randomly selecting the direction of a position from these occupied pixels as $\boldsymbol{d}_{m}$.
\textit{Average} means computing the average direction from these occupied pixels as $\boldsymbol{d}_{m}$.

\begin{equation}
\label{eq:prune}
    \boldsymbol{d}_{m} = \boldsymbol{d}_{j}, j =\min{\boldsymbol{A}_{\boldsymbol{p}_{i}}}
\end{equation}

\begin{figure*}[tb] \centering
    \includegraphics[width=0.8\textwidth]{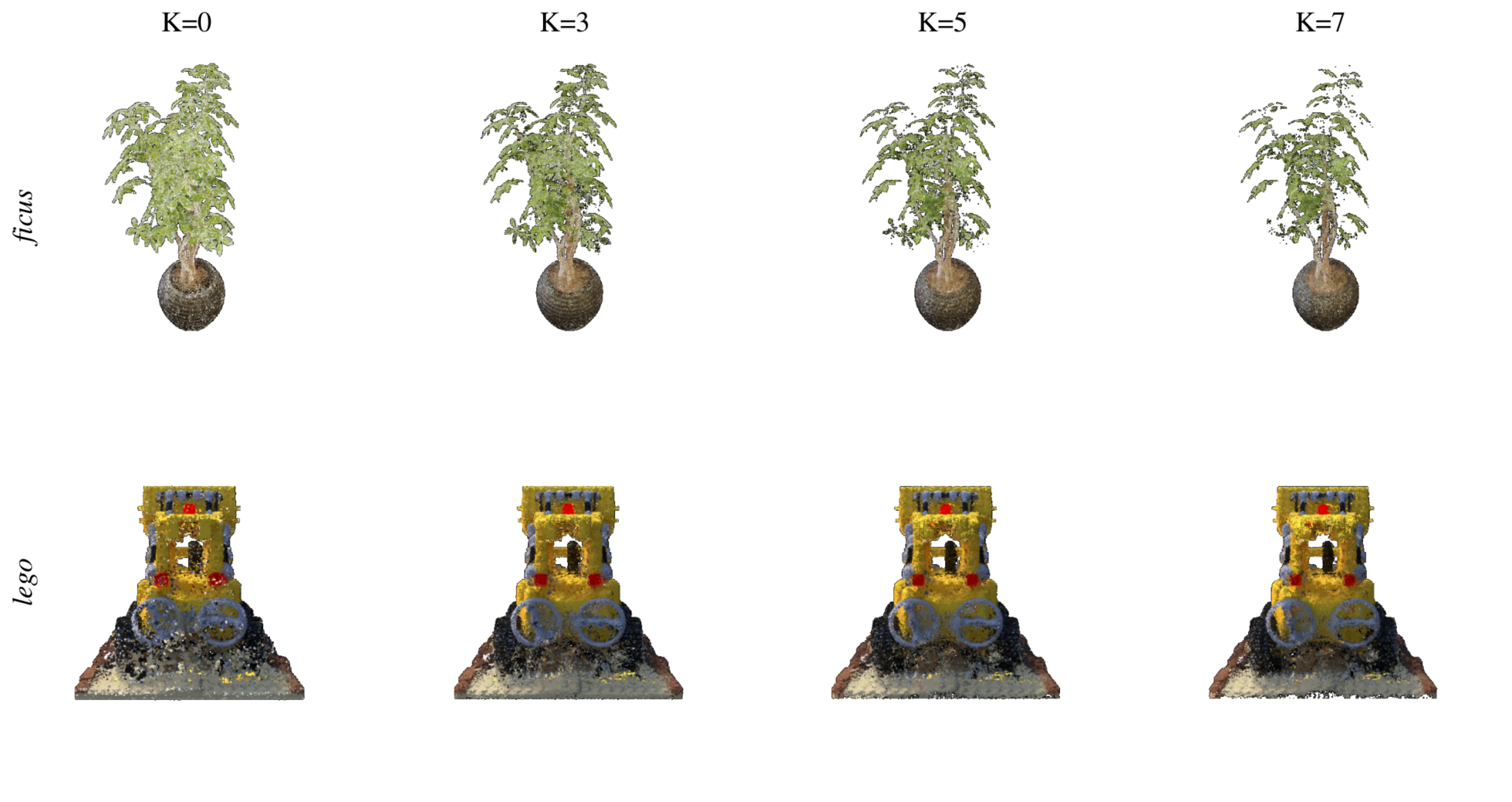}
    \caption{We assign colors to the z-buffers of different layers based on the point cloud to observe their geometric differences.}
    \label{fig:k_zbuffers}
\end{figure*}

\begin{figure*}[tb] \centering
    \includegraphics[width=0.9\textwidth]{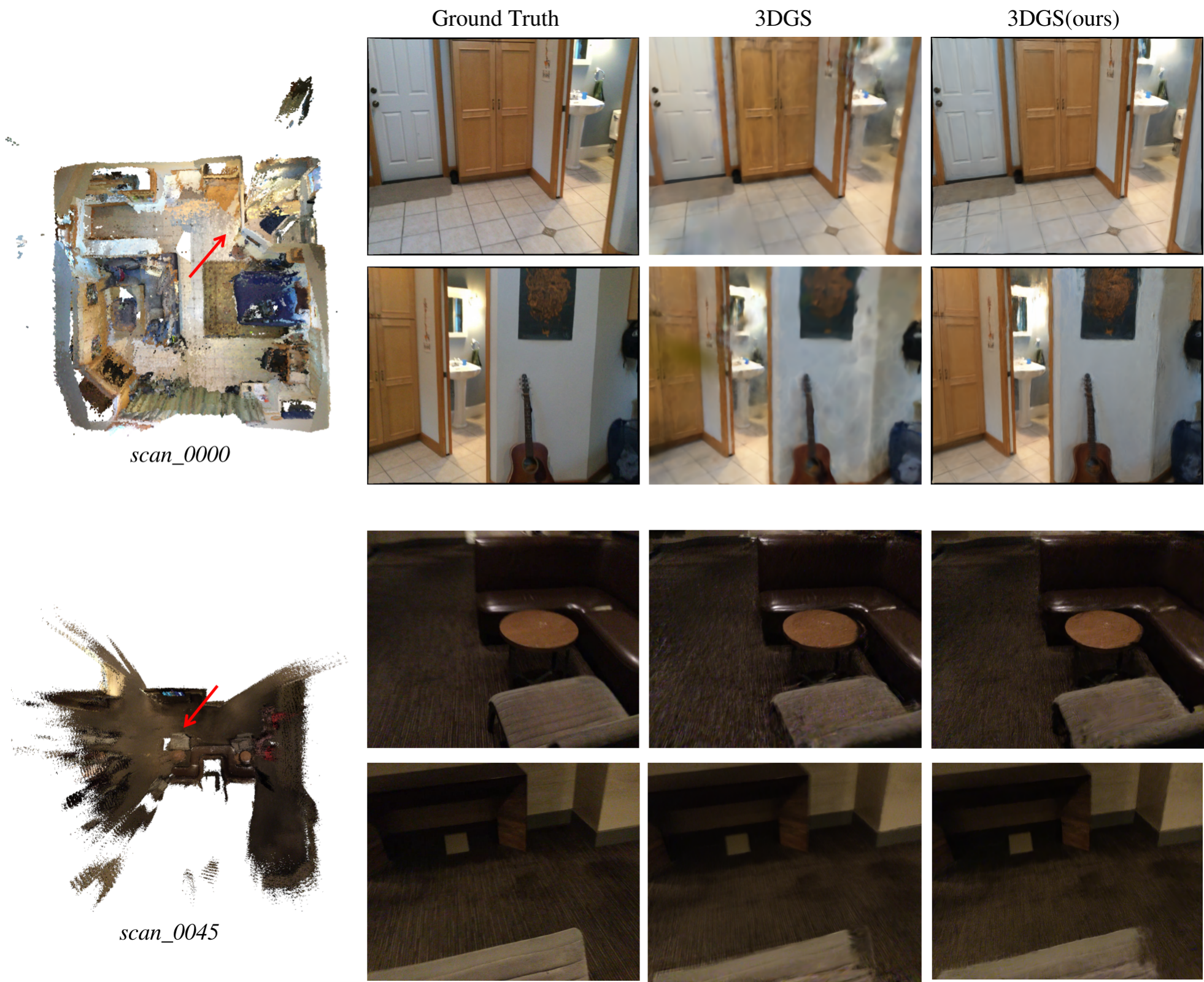}
    \caption{We assigned colors to the z-buffers of different layers based on the point cloud to observe their geometric differences.}
    \label{fig:vis_scannet}
\end{figure*}

\begin{figure*}[tb] \centering
    \includegraphics[width=\textwidth]{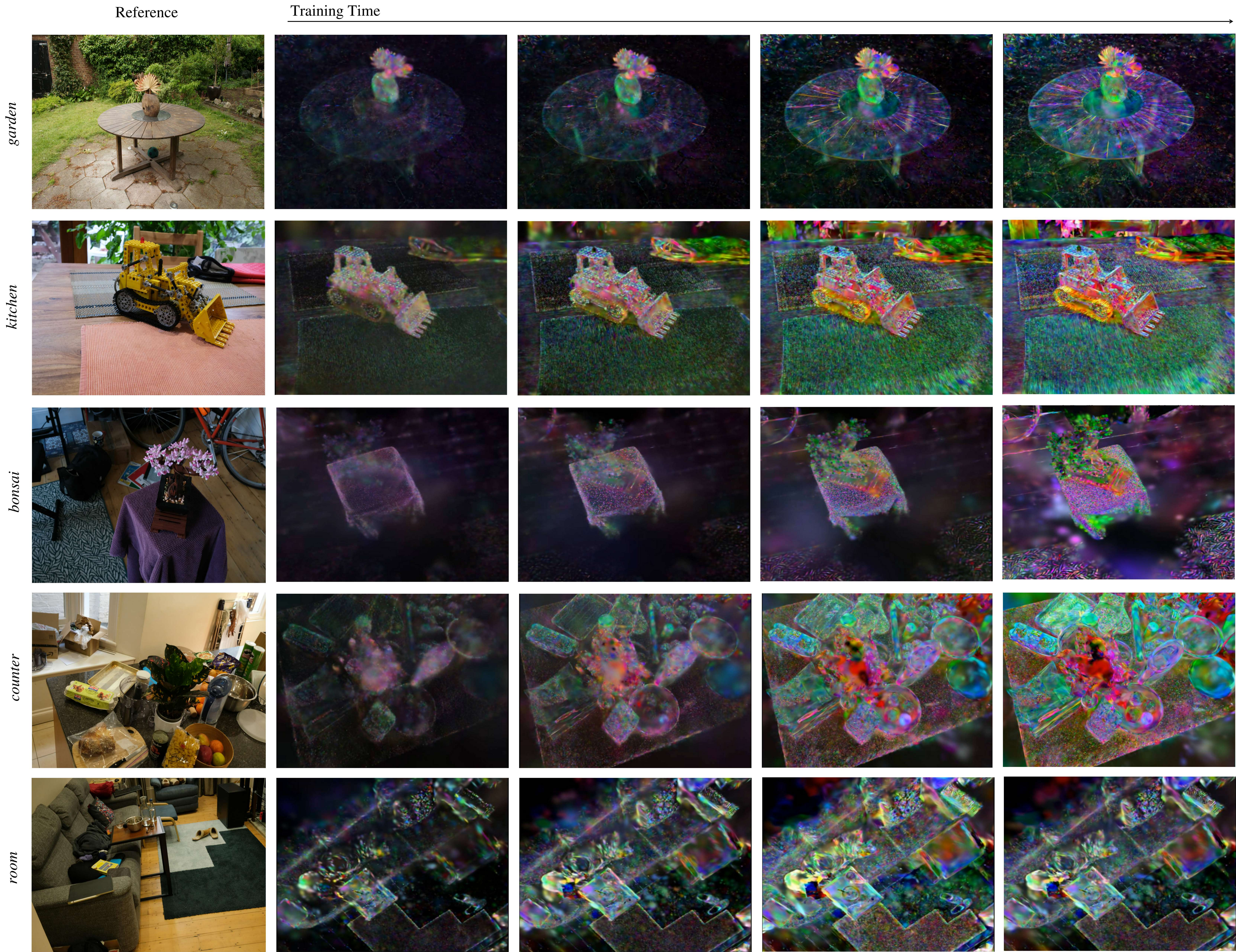}
    \caption{Radiance mapping changes over training time on Mip-NeRF dataset.}
    \label{fig:rm}
\end{figure*}

K z-buffers are visualized in Figure~\ref{fig:k_zbuffers}. 
The visual comparison of novel view synthesis results on the ScanNet dataset is presented in Figure~\ref{fig:vis_scannet}. We also visualize the changes in radiance mapping over time during the training process in Figure~\ref{fig:rm}.

\begin{table*}[!ht]
    \centering
    \setlength\tabcolsep{1pt}
    \caption{The impact of different components on rendering quality.}
    \vspace{-8pt}
    \begin{tabular*}{\hsize}{@{}@{\extracolsep{\fill}}lcccccccccc@{}}
    \toprule
     Data  & \multicolumn{3}{c}{NeRF-Synthetic} & \multicolumn{3}{c}{ScanNet} & \multicolumn{3}{c}{DTU} \\
    \cmidrule(r){2-4}\cmidrule(r){5-7} \cmidrule(r){8-10}
    \multicolumn{1}{l}{Metric} & PSNR$\uparrow$ & SSIM$\uparrow$ & LPIPS$\downarrow$ &  PSNR$\uparrow$ & SSIM$\uparrow$ & LPIPS$\downarrow$ & PSNR$\uparrow$ & SSIM$\uparrow$ & LPIPS$\downarrow$  \\
    \midrule
    BPCR &
        29.12 & 0.935 & 0.056 &
        25.88 & \textbf{0.794} & 0.415 &
        30.81 & 0.904 & \textbf{0.128} \\
    BPCR(KFN) &
        \textbf{29.95} & \textbf{0.940} & 0.055 &
        25.93 & 0.783 & 0.418 &
        \textbf{31.20} & \textbf{0.910} & 0.152 \\
    BPCR(KFN+Pruning) &
        29.68 & 0.937 & \textbf{0.054} &
        26.05 & 0.783 & 0.405 &
        31.08 & 0.906 & 0.153 \\
    BPCR(KFN+Pruning+Rect) &
        29.58 & 0.936 & \textbf{0.054} &
        \textbf{26.66} & 0.789 & \textbf{0.383} &
        30.99 & 0.907 & 0.151 & \\
    \midrule
    FrePCR &
    29.26 & 0.934 & 0.055 &
    \textbf{26.44} & \textbf{0.788} & \textbf{0.384} &
    24.61 & 0.870 & 0.195 & \\
    FrePCR(KFN) &
    29.44 & 0.935 & 0.058 &
    26.20 & \textbf{0.788} & 0.390 &
    29.20 & 0.889 & 0.167 & \\
    FrePCR(KFN+Pruning) &
    29.68 & 0.935 & 0.057 &
    26.12 & 0.785 & 0.404 &
    \textbf{30.78} & \textbf{0.905} & 0.156 & \\
    FrePCR(KFN+Pruning+Rect) &
    \textbf{29.83} & \textbf{0.939} & \textbf{0.051} &
    26.34 & \textbf{0.788} & 0.391 &
    30.61 & \textbf{0.905} & \textbf{0.153} &  \\
    \bottomrule
    \end{tabular*}
    \label{tab:benchmark}
\end{table*}

\begin{table*}[!ht]
    \centering
    \setlength\tabcolsep{1pt}
    \caption{PSNR$\uparrow$ on each scene in NeRF-Synthetic.}
    \vspace{-8pt}
    \begin{tabular*}{\hsize}{@{}@{\extracolsep{\fill}}lccccccccc@{}}
    \toprule
     Method & Chair & Drums & Ficus & Hotdog & Lego & Materials & Mic & Ship & Mean \\
    \midrule
    BPCR & \textbf{31.67} & 24.55 & 29.13 & \textbf{33.48} & 26.82 & \textbf{28.21} & \textbf{33.07} & 26.06 & 29.12 \\
    BPCR(ours) & 31.46 & \textbf{24.70} & \textbf{32.54} & 33.06 & \textbf{27.62} & 28.06 & 32.85 & \textbf{26.38} & \textbf{29.58} \\
    \midrule
    FrePCR & 31.68 & 24.54 & 29.55 & \textbf{33.86} & 26.74 & \textbf{28.81} & 32.91 & 25.97 & 29.26 \\
    FrePCR(ours) & \textbf{31.81} & \textbf{24.91} & \textbf{32.56} & 33.61 & \textbf{27.63} & 28.09 & \textbf{33.44} & \textbf{26.58} & \textbf{29.83} \\
    \midrule
    3DGS & \textbf{36.11} &	\textbf{26.15} &34.84 &	37.63 &	\textbf{36.13} & \textbf{30.16} & 35.37 & 30.58 & 33.37 \\
    3DGS(ours) & 36.01 & 25.63 & \textbf{36.71} & \textbf{38.33} & 35.80 & 29.90 & \textbf{35.84} & \textbf{30.79} & \textbf{33.62} \\
    \bottomrule
    \end{tabular*}
    \label{tab:benchmark}
\end{table*}

\begin{table*}[!ht]
    \centering
    \setlength\tabcolsep{1pt}
    \caption{SSIM$\uparrow$ on each scene in NeRF-Synthetic.}
    \vspace{-8pt}
    \begin{tabular*}{\hsize}{@{}@{\extracolsep{\fill}}lccccccccc@{}}
    \toprule
     Method & Chair & Drums & Ficus & Hotdog & Lego & Materials & Mic & Ship & Mean \\
    \midrule
    BPCR & \textbf{0.960} & 0.926 & 0.962 & \textbf{0.966} & 0.906 & \textbf{0.949} & \textbf{0.984} & \textbf{0.826} & 0.935 \\
    BPCR(ours) & 0.957 & \textbf{0.927} & \textbf{0.977} & 0.961 & \textbf{0.915} & 0.944 & 0.983 & 0.825 & \textbf{0.936} \\
    \midrule
    FrePCR & 0.959 & 0.926 & 0.964 & \textbf{0.964} & 0.903 & \textbf{0.951} & 0.983 & 0.821 & 0.934 \\
    FrePCR(ours) & \textbf{0.961} & \textbf{0.929} & \textbf{0.978} & \textbf{0.964} & \textbf{0.918} & 0.945 & \textbf{0.985} & \textbf{0.830} & \textbf{0.939} \\
    \midrule
    3DGS & \textbf{0.988} &	\textbf{0.954} &0.987 &	0.986 &	\textbf{0.984} &0.960 &	\textbf{0.992} &	0.896 &	0.968 \\
    3DGS(ours) & \textbf{0.988} & 0.949 & \textbf{0.991} & \textbf{0.987} & 0.983 & \textbf{0.962} & \textbf{0.992} & \textbf{0.898} & \textbf{0.969} \\
    \bottomrule
    \end{tabular*}
    \label{tab:benchmark}
\end{table*}

\begin{table*}[!ht]
    \centering
    \setlength\tabcolsep{1pt}
    \caption{LPIPS$\downarrow$ on each scene in NeRF-Synthetic.}
    \vspace{-8pt}
    \begin{tabular*}{\hsize}{@{}@{\extracolsep{\fill}}lccccccccc@{}}
    \toprule
     Method & Chair & Drums & Ficus & Hotdog & Lego & Materials & Mic & Ship & Mean \\
    \midrule 
    BPCR & 0.033 & 0.064 & 0.032 & \textbf{0.035} & 0.077 & \textbf{0.044} & \textbf{0.013} & 0.152 & 0.056 \\
    BPCR(ours) & \textbf{0.030} & \textbf{0.063} & \textbf{0.020} & 0.043 & \textbf{0.076} & 0.045 & 0.015 & \textbf{0.144} & \textbf{0.054} \\
    \midrule
    FrePCR & 0.030 & 0.064 & 0.027 & \textbf{0.035} & 0.079 & \textbf{0.044} & 0.014 & 0.145 & 0.055 \\
    FrePCR(ours) & \textbf{0.028} & \textbf{0.061} & \textbf{0.019} & 0.036 & \textbf{0.068} & 0.045 & \textbf{0.012} & \textbf{0.140} & \textbf{0.051} \\
    \midrule
    3DGS & 0.011 & 0.037 & 0.012 & 0.017 & 0.013 & 0.031 & 0.006 & 0.095 & 0.028 \\
    3DGS(ours) & \textbf{0.007} & \textbf{0.034} & \textbf{0.005} & \textbf{0.007} & \textbf{0.008} & \textbf{0.017} & \textbf{0.004} & \textbf{0.068} & \textbf{0.019} \\
    \bottomrule
    \end{tabular*}
    \label{tab:benchmark}
\end{table*}

\begin{table*}[!ht]
    \centering
    \setlength\tabcolsep{1pt}
    \caption{PSNR$\uparrow$ on each scene in ScanNet and DTU.}
    \vspace{-8pt}
    \begin{tabular*}{\hsize}{@{}@{\extracolsep{\fill}}lcccccccc@{}}
    \toprule
     Method & ScanNet-0000 & ScanNet-0043 & ScanNet-0045 & Mean & DTU-110 & DTU-114 & DTU-118 & Mean \\
    \midrule
    BPCR & 23.79 & 25.26 & 28.59 & 25.88 & \textbf{30.79} & 27.91 & 33.72 & 30.81 \\
    BPCR(ours) & \textbf{24.58} & \textbf{26.20} & \textbf{29.21} & \textbf{26.66} & 30.63 & \textbf{28.30} & \textbf{34.03} & \textbf{30.99} \\
    \midrule
    FrePCR & 24.03 & \textbf{26.36} & \textbf{28.92} & \textbf{26.44} & 22.29 & 25.87 & 25.66 & 24.61 \\
    FrePCR(ours) & \textbf{24.32} & 25.82 & 28.88 & 26.34 & \textbf{30.22} & \textbf{28.40} & \textbf{33.20} & \textbf{30.61} \\
    \midrule
    3DGS & 18.54 & \textbf{25.61} & \textbf{28.34} & 24.16 & 33.43 & \textbf{30.60} & 37.69 & \textbf{33.91} \\
    3DGS(ours) & \textbf{23.74} & 24.62 & 28.29 & \textbf{25.55} & \textbf{33.56} & 30.10 & \textbf{38.07} & \textbf{33.91} \\
    \bottomrule
    \end{tabular*}
    \label{tab:benchmark}
\end{table*}

\begin{table*}[!ht]
    \centering
    \setlength\tabcolsep{1pt}
    \caption{SSIM$\uparrow$ on each scene in ScanNet and DTU.}
    \vspace{-8pt}
    \begin{tabular*}{\hsize}{@{}@{\extracolsep{\fill}}lcccccccc@{}}
    \toprule
     Method & ScanNet-0000 & ScanNet-0043 & ScanNet-0045 & Mean & DTU-110 & DTU-114 & DTU-118 & Mean \\
    \midrule
    BPCR & \textbf{0.748} & 0.844 & \textbf{0.789} & \textbf{0.794} & \textbf{0.924} & 0.863 & 0.926 & 0.904 \\
    BPCR(ours) & 0.742 & \textbf{0.845} & 0.781 & 0.789 & 0.923 & \textbf{0.871} & \textbf{0.927} & \textbf{0.907} \\
    \midrule
    FrePCR & \textbf{0.744} & 0.841 & \textbf{0.779} & \textbf{0.788} & 0.870 & 0.846 & 0.893 & 0.870 \\
    FrePCR(ours) & 0.742 & \textbf{0.842} & 0.778 & \textbf{0.788} & \textbf{0.921} & \textbf{0.873} & \textbf{0.922} & \textbf{0.905} \\
    \midrule
    3DGS & 0.731 & \textbf{0.837} & 0.751 & 0.773 & 0.951 & 0.920 & 0.960 & 0.943 \\
    3DGS(ours) & \textbf{0.763} & 0.833 & \textbf{0.760} & \textbf{0.785} & \textbf{0.958} & \textbf{0.935} & \textbf{0.970} & \textbf{0.954} \\
    \bottomrule
    \end{tabular*}
    \label{tab:benchmark}
\end{table*}

\begin{table*}[!ht]
    \centering
    \setlength\tabcolsep{1pt}
    \caption{LPIPS$\downarrow$ on each scene in ScanNet and DTU.}
    \vspace{-8pt}
    \begin{tabular*}{\hsize}{@{}@{\extracolsep{\fill}}lcccccccc@{}}
    \toprule
     Method & ScanNet-0000 & ScanNet-0043 & ScanNet-0045 & Mean & DTU-110 & DTU-114 & DTU-118 & Mean \\
    \midrule
    BPCR & 0.440 & 0.389 & 0.415 & 0.415 & 0.114 & \textbf{0.154} & 0.117 & \textbf{0.128} \\
    BPCR(ours) & \textbf{0.415} & \textbf{0.349} & \textbf{0.387} & \textbf{0.383} & \textbf{0.108} & 0.233 & \textbf{0.111} & 0.151 \\
    \midrule
    FrePCR & \textbf{0.400} & \textbf{0.352} & \textbf{0.400} & \textbf{0.384} & 0.168 & 0.264 & 0.154 & 0.195 \\
    FrePCR(ours) & 0.417 & 0.356 & \textbf{0.400} & 0.391 & \textbf{0.109} & \textbf{0.235} & \textbf{0.116} & \textbf{0.153} \\
    \midrule
    3DGS & 0.496 & \textbf{0.330} & \textbf{0.382} & \textbf{0.403} & \textbf{0.046} & \textbf{0.063} & \textbf{0.040} & \textbf{0.050} \\
    3DGS(ours) & \textbf{0.440} & 0.359 & 0.420 & 0.406 & 0.047 & 0.076 & 0.054 & 0.059 \\
    \bottomrule
    \end{tabular*}
    \label{tab:benchmark}
\end{table*}

\begin{table*}[!ht]
    \centering
    \setlength\tabcolsep{1pt}
    \caption{PSNR$\uparrow$ on each scene in Mip-NeRF360.}
    \vspace{-8pt}
    \begin{tabular*}{\hsize}{@{}@{\extracolsep{\fill}}lcccccccc@{}}
    \toprule
     Method & bicycle & bonsai & counter & garden & kitchen & room & stump & Mean \\
     \midrule
     InstantNGP & 23.00 & 30.30 & 26.56 & 24.78 & 29.02 & 29.16 & 23.84 & 26.67 \\
     Mip-Splatting & 25.25 & 31.96 & 29.04 & 27.47 & 31.25 & 31.54 & 26.64 & 29.02 \\
    \midrule
    3DGS & \textbf{25.15} & 32.07 & 28.99 & 27.25 & 31.16 & 31.30 &\textbf{26.57} & 28.93 \\
    3DGS(ours) & 24.27 & \textbf{32.49} & \textbf{29.54} & \textbf{27.36} & \textbf{32.51} & \textbf{32.66} & 25.49 & \textbf{29.19} \\
    \bottomrule
    \end{tabular*}
    \label{tab:benchmark}
\end{table*}

\begin{table*}[!ht]
    \centering
    \setlength\tabcolsep{1pt}
    \caption{SSIM$\uparrow$ on each scene in Mip-NeRF360.}
    \vspace{-8pt}
    \begin{tabular*}{\hsize}{@{}@{\extracolsep{\fill}}lcccccccc@{}}
    \toprule
     Method & bicycle & bonsai & counter & garden & kitchen & room & stump & Mean \\
    \midrule
    InstantNGP & 0.526 & 0.890 & 0.812 & 0.654 & 0.844 & 0.850 & 0.590 & 0.738 \\
    Mip-Splatting & 0.765 & 0.941 & 0.907 & 0.869 & 0.926 & 0.918 & 0.774 & 0.871\\
    \midrule
    3DGS & \textbf{0.763} & 0.940 & 0.906 & \textbf{0.864} & 0.926 & 0.917 & \textbf{0.770} & \textbf{0.869} \\
    3DGS(ours) & 0.727 & \textbf{0.942} & \textbf{0.909} & 0.859 & \textbf{0.935} & \textbf{0.926} & 0.714 & 0.859 \\
    \bottomrule
    \end{tabular*}
    \label{tab:benchmark}
\end{table*}

\begin{table*}[!ht]
    \centering
    \setlength\tabcolsep{1pt}
    \caption{LPIPS$\downarrow$ on each scene in Mip-NeRF360.}
    \vspace{-8pt}
    \begin{tabular*}{\hsize}{@{}@{\extracolsep{\fill}}lcccccccc@{}}
    \toprule
     Method & bicycle & bonsai & counter & garden & kitchen & room & stump & Mean \\
     \midrule
     InstantNGP & 0.489 & 0.295 & 0.373 & 0.346 & 0.255 & 0.383 & 0.439 & 0.369 \\
     Mip-Splatting & 0.243 & 0.254 & 0.258 & 0.124 & 0.155 & 0.286 & 0.251 & 0.224 \\
    \midrule
    3DGS & 0.173 & 0.134 & 0.156 & 0.076 & 0.095 & 0.166 & \textbf{0.153} & 0.136 \\
    3DGS(ours) & \textbf{0.153} & \textbf{0.125} & \textbf{0.144} & \textbf{0.064} & \textbf{0.084} & \textbf{0.149} & 0.161 & \textbf{0.126} \\
    \bottomrule
    \end{tabular*}
    \label{tab:benchmark}
\end{table*}

\begin{table*}[!ht]
    \centering
    \setlength\tabcolsep{1pt}
    \caption{Storage(MB)$\downarrow$ on each scene in NeRF-Synthetic.}
    \vspace{-8pt}
    \begin{tabular*}{\hsize}{@{}@{\extracolsep{\fill}}lccccccccc@{}}
    \toprule
     Method & Chair & Drums & Ficus & Hotdog & Lego & Materials & Mic & Ship & Mean \\
    \midrule
    3DGS & 125 & 135 & 90 & 156 & 197 & 213 & 113 & 335 & 170.50 \\
    3DGS(ours) & \textbf{39} & \textbf{36} & \textbf{24} & \textbf{29} & \textbf{36} & \textbf{30} & \textbf{31} & \textbf{54} & \textbf{34.87} \\
    \bottomrule
    \end{tabular*}
    \label{tab:benchmark}
\end{table*}

\begin{table*}[!ht]
    \centering
    \setlength\tabcolsep{1pt}
    \caption{Storage(MB)$\downarrow$ on each scene in ScanNet and DTU.}
    \vspace{-8pt}
    \begin{tabular*}{\hsize}{@{}@{\extracolsep{\fill}}lcccccccc@{}}
    \toprule
     Method & ScanNet-0000 & ScanNet-0043 & ScanNet-0045 & Mean & DTU-110 & DTU-114 & DTU-118 & Mean \\
    \midrule
    3DGS & 694 & 486 & 438 & 539.33 & 250 & 339 & 259 & 282.66 \\
    3DGS(ours) & \textbf{221} & \textbf{75} & \textbf{92} & \textbf{129.33} & \textbf{45} & \textbf{55} & \textbf{36} & \textbf{45.33} \\
    \bottomrule
    \end{tabular*}
    \label{tab:benchmark}
\end{table*}

\begin{table*}[!ht]
    \centering
    \setlength\tabcolsep{1pt}
    \caption{Storage(MB)$\downarrow$ on each scene in Mip-NeRF360.}
    \vspace{-8pt}
    \begin{tabular*}{\hsize}{@{}@{\extracolsep{\fill}}lcccccccc@{}}
    \toprule
     Method & bicycle & bonsai & counter & garden & kitchen & room & stump & Mean \\
    \midrule
    InstantNGP & 41 & 37 & 37 & 38 & 37 & 39 & 41 & 38.57 \\
    Mip-Splatting & 1521 & 313 & 299 & 1212 & 448 & 394 & 1195 & 768.86 \\
    \midrule
    3DGS & 1419 & 272 & 265 & 1319 & 403 & 355 & 1116 & 735.57 \\
    3DGS(ours) & \textbf{880} & \textbf{168} & \textbf{167} & \textbf{583} & \textbf{192} & \textbf{207} & \textbf{487} & \textbf{383.43} \\
    \bottomrule
    \end{tabular*}
    \label{tab:benchmark}
\end{table*}

\clearpage
\newpage



\section*{Acknowledgments}
This work was supported by the Key R\&D Program of Zhejiang under Grant (2025C03001), the Fundamental Research Funds for the Provincial Universities of Zhejiang (GK259909299001-023), the National Nature Science Foundation of China (62301198).


\bibliographystyle{named}
\bibliography{ijcai25}

\end{document}